\documentclass[letterpaper, 10 pt, conference]{ieeeconf}  

\IEEEoverridecommandlockouts                              

\usepackage{graphicx}
\usepackage{epsfig} 
\usepackage{amsmath} 
\usepackage{mathtools}
\usepackage{bm}
\usepackage{amssymb}  
\usepackage{subfigure}
\usepackage{algorithm}
\usepackage[noend]{algpseudocode}
\usepackage{multirow}
\usepackage{booktabs}
\usepackage{color}

\usepackage{soul}
\newcommand{\luisrm}[1]{}

\newcommand{\forcef}{f}

\newcommand{\myvec}[1]{\boldsymbol{{#1}}}
\newcommand{\mymatrix}[1]{\boldsymbol{{#1}}}
\newcommand{\human}{\text{h}}
\newcommand{\obj}{\text{o}}
\newcommand{\tool}{\text{t}}

\title{\LARGE \bf
    Planning for Muscular and Peripersonal-Space Comfort \\ 
    during Human-Robot Forceful Collaboration
}

\author{Lipeng Chen$^*$, Luis F C Figueredo$^*$, Mehmet R. Dogar
\thanks{$^*$First two authors contributed equally to this work.
  \newline  Authors are with School of Computing, University of Leeds, Leeds, UK,
        {\tt\small \{sclc,l.figueredo,m.r.dogar\}@leeds.ac.uk}}   
\thanks{This project has received funding from the European Union's Horizon 2020 research and innovation programme under the Marie Sklodowska-Curie grants agreement No. 746143 and 795714, and from the UK Engineering and Physical Sciences Research Council under grant EP/P019560/1.
}
}

\begin{document}

\maketitle
\thispagestyle{empty}
\pagestyle{empty}

\begin{abstract}

\textit{This work has been accepted and will appear in the 2018 IEEE-RAS International Conference on Humanoid Robots (HUMANOID 2018).}

This paper presents a planning algorithm designed to improve cooperative robot behaviour concerning human comfort during forceful human-robot physical interaction. Particularly,  we are interested in planning for object grasping and positioning ensuring not only stability against the exerted human force but also empowering the robot with capabilities to address and improve human experience and comfort. 
Herein, comfort is addressed as both the muscular activation level required to exert the cooperative task, 
and the human spatial perception during the interaction, namely, the peripersonal space. 
By maximizing both comfort criteria, the robotic system 
can plan for the task (ensuring grasp stability) \textit{and} for the human (improving human comfort). 
We believe this to be a key element to achieve intuitive and fluid human-robot interaction in real applications. 
Real HRI drilling and cutting experiments 
illustrated the efficiency of the proposed planner in improving overall comfort and HRI experience without compromising grasp stability.
%

\end{abstract}

\section{Introduction}
\label{sec:intro} 

\begin{figure}[!t]
	\begin{center}
		\mbox{
			
			\subfigure[Too close to the robot]{{
					\label{fig:subfig:unsafesafe}  
					\includegraphics[width=1.6 in, angle=-0]{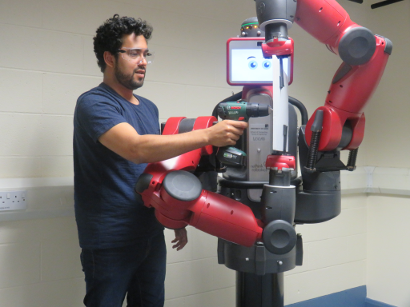}}}
			
			\subfigure[High muscular effort]{{
					\label{fig:subfig:uncomfort}  
					\includegraphics[width=1.6  in, angle=-0]{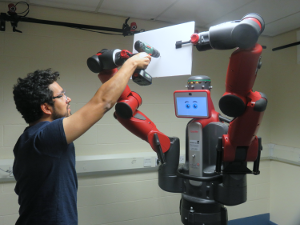}}}			
		}
		\mbox{
	
			\subfigure[Unstable grasp]{{
					\label{fig:subfig:uncomfortSafe}  
					\includegraphics[width=1.6 in, angle=-0]{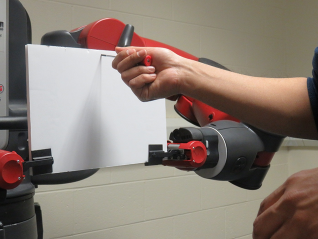}}}
			
			\subfigure[Comfortable and stable]{{
					\label{fig:subfig:comfortsafe}  
					\includegraphics[width=1.6  in, angle=-0]{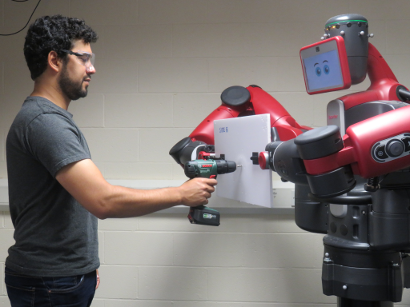}}}			
		}
		\caption{Human-robot collaborative drilling on a board.}
		\label{fig:h-r drilling}
	\end{center}
\end{figure}

In this paper, we are interested in the problem of a humanoid robot
manipulating an object in the close proximity of a human operator, who applies
a forceful operation such as drilling and cutting on the object. An instance is
shown in Fig.~\ref{fig:h-r drilling} where a human is drilling on a wooden
board firmly held by a humanoid robot at different configurations.
Theoretically, there may be infinite ways for the robot to grasp and position
the board and accordingly for the human to apply the desired force on it, among
which a large proportion of configurations are obviously uncomfortable for the human as shown
in Fig.~\ref{fig:subfig:uncomfort} and even unsafe as shown in Fig.~\ref{fig:subfig:unsafesafe}.
Human comfort and safety
during a forceful operation highly depends on how the human body is positioned and configured,
and therefore, where and how the robot positions its manipulators and the
object for the operation.  Under this circumstance, given a forceful operation to be applied by a
human operator, it is critical to empower the robot with the capability to plan
and position itself and the object at configurations which are not only stable,
but also comfortable for the human to perform the operation.  

In this context,
we propose a planner that enables a robot to grasp and position an object for a
forceful operation applied by a human that explicitly concerns human comfort
and force stability.  The final goal is to have an enhanced human-robot
interaction (HRI) experience with improved performance from both sides.

%
%


To quantify the human comfort during forceful interaction, we formulate a cost
based on human muscular features, kinematics and the involved forces.  First,
our planner produces 
solutions 
by minimizing a human \textit{muscular
comfort} cost, which quantifies the muscular
effort required for the human to perform a specific forceful operation on the
object.  
This metric, for example, predicts higher muscular comfort for the human 
in Fig.~\ref{fig:subfig:comfortsafe} and Fig.~\ref{fig:subfig:unsafesafe}, and lower muscular comfort for the human in Fig.~\ref{fig:subfig:uncomfort}, for the same task of drilling.
The muscular comfort metric predicts and
proactively instructs human to configurations which require less physical
joint-torque efforts to perform a forceful operation.

Nonetheless, through our initial experiments, we discovered that optimizing
only the muscular comfort is not sufficient. For example, while the human in Fig.~\ref{fig:subfig:unsafesafe} may have better muscular comfort compared to Fig.~\ref{fig:subfig:uncomfort}, he is dangerously close to the robot and is obstructed by the robot links.
Human behavior vary according to their assessment of the robot counterpart, and
human performance also depends on their level of trust and safety perception regarding the robot coworker \cite{2014_Lasota_Rossano_Shah__CASE,2015_Lasota_Shah__HF}. 
Furthermore, the human and robot may need to move before, during and after the forceful operation. Therefore, having enough space between them is important.
In this paper, to improve human awareness and safety perception, we exploit the concept of peripersonal space comfort. The idea is to allow humans to move and act within their peripersonal space%
  \footnote{Peripersonal space is the space immediately surrounding our body, or the sector of space that closely surrounds a certain body part, in which multisensory and sensorimotor integration is enhanced \cite{2014_Bartolo_Carlier_Hassaini_Martin_Coello___FrontiersHN,1981_Rizzolatti_Scandolara_Matelli_Gentilucci___BehaviouralBrainResearch}. } 
  minimizing the perceived risk of robot intervention. Therefore, we propose a \textit{peripersonal-space cost} that is based on the distance between parts of the human and robot bodies.    
%
The configurations in Fig.~\ref{fig:subfig:comfortsafe} shows a configuration optimizing for both the muscular and the peripersonal-space comfort: the human is drilling at a relatively comfortable pose, with the distance to the robot being large enough to reduce spatial discomfort.

Finally, ensuring robot capabilities to cope with the forceful task is also crucial, and cannot be decoupled from system analysis since different object/human poses results in different forces to cope. 
Force stability checking involves choosing appropriate robot grasps on an
object and joint configurations to hold the object stable against the force
applied by the human. 
A bad grasp of the object (as in Fig.~\ref{fig:subfig:uncomfortSafe}) may result in failed operations which would not only disturb the
fluency of HRI, but also pose serious danger to human
safety. For example, the object may slip through the fingers during a cutting
operation, or it may bend away
from the desired pose due to large torques around the gripper during a drilling
action as shown in Fig.~\ref{fig:subfig:uncomfortSafe}.
This places an additional constraint: tightly coupled
human-robot kinematics and force stability must be guaranteed, while the human
comfort is optimized in the resulting space. 

In our previous work \cite{chen2018manipulation}, we have presented a manipulation planner for a robot to keep an object stable under changing external forces by
automatically deciding when and how to regrasp in a multi-step task scenario. 
Although improved robot performance has been achieved, human comfort during the interaction was ignored. To be more specific, the object pose was fixed in that work, which posed difficulties and discomfort for the human to perform the task. In real world applications, the resulting grasp/object pose may be crucial for the human experience, performance and safety. 
Different from our previous work, the work in this paper focuses on planning
robot and object configurations that are not only stable but also comfortable
for the human to perform the forceful operation. It can be regarded as the
first step of planning fluid, comfortable and efficient human-robot
collaborative forceful manipulation. 

\section{Related work}
\label{sec:related}

Human-robot collaboration (HRC) has been interpreted from various aspects and tackled at different levels. 
Existing work in forceful HRC mostly addresses the control problem \cite{kosuge1997control,rozo2016learning}, solving for the necessary stiffness of manipulator joints as an external force is applied \cite{gopinathan2017user}, and assumes the object to be already grasped at pre-specified positions by the robot. 
Similar planning work mostly focuses on handover  \cite{sisbot2012human,strabala2013towards}, transportation \cite{rozo2016learning,solanes2018human}, or scenarios where the robot avoids colliding a human in the same workspace \cite{luo2018unsupervised,maeda2017probabilistic}. 
Handover resembles our task in requiring the robot to position an object for a human. 
However, the physical interactions for handovers are usually 
not as intensive or physical demanding for both parts
thus planning has other priorities 
\cite {sisbot2012human,parastegari2017modeling}. 
Our task scenario differs in the existence of strong forceful interactions. 
Similar to handover, human-robot collaborative transportation connects a human and a robot together through a shared object 
but for a longer time. 
It requires the human and the robot to share load by holding an object at a desired orientation \cite{lawitzky2010load}, which is similar to finding appropriate grasps and robot configurations to keep an object stable in our work. 
Human comfort and safety have been explicitly considered in human-aware motion planning for 
for object handover and transportation \cite{sisbot2012human,parastegari2017modeling,solanes2018human}. 
With regard to forceful applications, Mansfeld et al. \cite{mansfeld2018safety} extend the safety analysis to general HRI and propose a new concept global safety assessment framework in practical HRC  applications, namely the safety map. 
The excel works of Peternel et al. \cite {peternel2016adaptation,2017_Peternel_Kim_Ajoudani__Humanoids} should also be mentioned for addressing important topics in forceful HRC 
related to this work such as human ergonomics during forceful control co-operations and hand-over, where joint torques are taken into account through an offline identification technique but, still without proper knowledge on muscular activity---as well as safety perception and grasp stability.   
Similar problem has also been addressed in \cite{2018_ICRA_Busch_Toussaint_Lopes} 
but considering an industrial ergonomic assessment based on payload mass and kinematics alone.  
%
%
%
%
Indeed, despite the advances and contributions aforementioned, HRC still lacks  
efficient metrics for planning comfortable human-robot forceful collaborations 
explicitly considering muscular activation levels and human comfort perception, and planning object grasps and positioning. 
%
%

\section{Problem formulation}
\label{sec:defi} 

We are interested in the problem of finding optimal robot and object configurations for forceful human-robot collaborative manipulation, in which a human performs a specific forceful operation such as drilling and cutting, on an object stably held by a robot in the shared environment.

\begin{figure}[!t]
	\begin{center}
		\mbox{
			\subfigure{{
					\label{fig:subfig:problem_define}  
					\includegraphics[width=3.0 in,height=2.5 in, angle=-0]{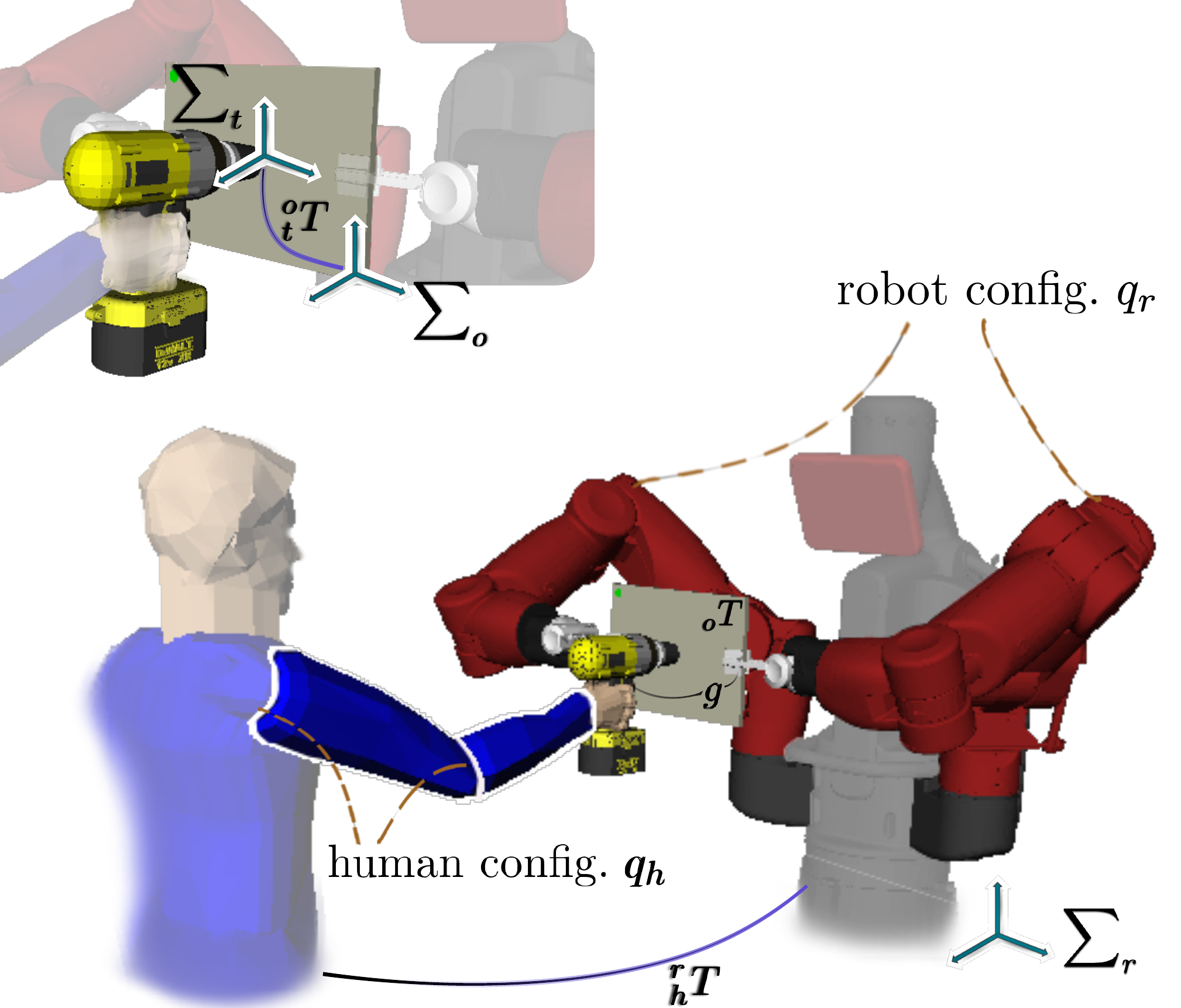}}}			
		}
		
		\caption{human-tool-object-robot configuration.}
		\label{fig:problem_define}
	\end{center}
\end{figure}

As shown in Fig.~\ref{fig:problem_define}, when the human applies the force, the human and robot geometry and kinematics are tightly coupled by means of the tool-object system. We assume a  fixed grasp  for the human to grasp a specific tool, and therefore a fixed kinematic transformation between the human hand and the tool. 
We assume a tool frame ${\sum}_{\text{t}}$ attached at the tooltip and a human frame attached at the shoulder ${\sum}_{\text{h}}$. Since the operational force is applied exactly at the contact points between the object and the tooltip, the force position (e.g., the point to drill on the object) can be specified by the pose  of the tooltip $_{\text{t}}^{\text{o}}\textbf{\textit{T}}$ in the object\footnote{We use superscripts at the top-left of the variables to distinguish the coordinate frames with respect to which the variables are expressed.} frame ${\sum}_{\text{o}}$. 

In this context, a forceful operation can be specified by a tuple: \begin{equation}\label{key}
(^\text{o}\textbf{\textit{f}},\, _{\text{t}}^{\text{o}}{\textbf{\textit{T}}})
\end{equation}
where $^\text{o}\textbf{\textit{f}}$ is the 6D generalized force vector.
We assume that such an operation is provided to our system; i.e. the pose of the force application (e.g. the drill point on the board) and the expected force (e.g. the expected drilling force).

Besides, we use a humanoid robot in our experiments, thus assuming the robot has two dexterous manipulators and each manipulator is equipped with a parallel gripper. However, our formulation can be easily extended to systems with more manipulators. Given a forceful operation $(^\text{o}\textbf{\textit{f}},\, _{\text{t}}^{\text{o}}{\textbf{\textit{T}}})$, the composite configuration for the human robot collaboration can be fully specified by five variables\footnote{Without loss of generality, we assume the robot frame ${\sum}_{\text{r}}$ coincides with the world frame, thus ommitting superscripts for variables if they are defined in the world/robot frame.} ${\textbf{\textit{q}}=\left(\textbf{\textit{q}}_{\text{h}},\,\textbf{\textit{q}}_{\text{r}},\,_{\text{h}}\textbf{\textit{T}},\,_{\text{o}}\textbf{\textit{T}},\textit{\textbf{g}}\right)}$, where:
\begin{itemize}
    \item $\textbf{\textit{q}}_{\text{h}}$ refers to human body configuration\footnote{In this work, the muscular comfort metric (see Sec.~\ref{sec:Muscular}) only takes human arm configuration into account, and therefore within this paper we use $q_{\text{h}}$ to refer to human arm configuration.  However, the optimization problem we define in this section is general to whole body configurations.};
	\item $\textbf{\textit{q}}_{\text{r}}=\left\{\textbf{\textit{q}}_{\text{rl}},\,\textbf{\textit{q}}_{\text{rr}}\right\}$ refers to the robot configuration, where $\textbf{\textit{q}}_{\text{rl}}$ and $\textbf{\textit{q}}_{\text{rr}}$ denote the left and right arm configuration respectively;
	\item $_{\text{h}}^{\text{}}\textbf{\textit{T}}$ describes the 6D pose of the human body in the robot frame;
	\item  $_{\text{o}}\textbf{\textit{T}}$ refers to the 6D object pose in the robot frame;
	\item $\textit{\textbf{g}}$ denotes a grasp using the pose of gripper(s)  with respect to the object. To be more specific, in this paper, $\textit{\textbf{g}}$ refers to a bimanual grasp specifying the pose of both left and right grippers on the object. 	
\end{itemize}
Note that there is redundancy in this definition. Specifically, given a robot configuration $\textbf{\textit{q}}_{\text{r}}$ and object pose $_{\text{o}}\textbf{\textit{T}}$, the corresponding grasp comfiguration $\textit{\textbf{g}}$ can be computed via robot forward kinematics.

\subsection{Optimization}

For a given force operation $(^\text{o}\textbf{\textit{f}},\, _{\text{t}}^{\text{o}}{\textbf{\textit{T}}})$, our planner aims at finding the robot configuration $\textbf{\textit{q}}_{\text{r}}$ and object pose  $_{\text{o}}\textbf{\textit{T}}$, such that the object is stably grasped against the applied force \textit{and} the human comfort is maximized:
\begin{equation}\label{optimization}
\begin{split}
&\textbf{\textit{q}}_{\text{r}}^*,\,_{\text{o}}\textbf{\textit{T}}^*=
\mathop{\arg\max}_{\textbf{\textit{q}}_{\text{r}},\,_{\text{o}}\textbf{\textit{T}}}\,\text{Comfort}\left(\textbf{\textit{q}}_{\text{r}},\,_{\text{o}}\textbf{\textit{T}},\,^\text{o}\textbf{\textit{f}},\, _{\text{t}}^{\text{o}}{\textbf{\textit{T}}}\right)\\
&\text{s.t.\,}\,\,\text{Is\_stable}\left(\textbf{\textit{q}}_{\text{r}},\,\textit{\textbf{g}},\,^\text{o}\textbf{\textit{f}},\, _{\text{t}}^{\text{o}}{\textbf{\textit{T}}}\right)
\end{split}
\end{equation}

Function $\text{Is\_stable}$ checks whether the robot at the configuration $\textbf{\textit{q}}_{\text{r}}$ is able to resist the operation force $^\text{o}\textbf{\textit{f}}$ with the grasp $\textit{\textbf{g}}$ on the object via static equilibrium, which will be discussed with more detail in Sec.~\ref{sec:Feasibility}. 

The optimization problem requires us to compute the value of $\text{Comfort}$,
which we use to represent the total human comfort. We propose to
model this value using two components: human's \textit{muscular} comfort and
human's \textit{peripersonal-space} comfort.
Here we assume the existence of functions which quantify these comfort metrics: 
\begin{itemize}
    \item $\text{Muscular}\left(\textbf{\textit{q}}_{\text{h}}, \, _{\text{h}}{\textbf{\textit{T}}},  \, ^{\text{o}}{\textbf{\textit{f}}}, \, _{\text{t}}^{\text{o}}{\textbf{\textit{T}}}, \, _{\text{o}}{\textbf{\textit{T}}} \right)$: This function
    returns a scalar value quantifying the muscular comfort of the human given
    the human configuration, pose, and the forceful operation to be applied. We
    present how we model this function in Sec.~\ref{sec:Muscular}.  
    \item $\text{Peripersonal}\left(\textbf{\textit{q}}_{\text{h}},\,_{\text{h}}{\textbf{\textit{T}}},\,\textbf{\textit{q}}_{\text{r}}\right)$:
        This function returns a scalar value quantifying the peripersonal-space
        comfort of the human given the human configuration, pose, and the robot
        configuration.  We present how we model this function in Sec.~\ref{sec:Peripersonal}.  
\end{itemize}

For a given forceful operation, our goal is to determine a robot configuration $\textbf{\textit{q}}_{\text{r}}$ and object pose $_{\text{o}}\textbf{\textit{T}}$ that maximizes these two costs.
However, computing these comfort metrics requires us to know the human configuration $\textbf{\textit{q}}_{\text{h}}$.
But, for a given robot configuration $\textbf{\textit{q}}_{\text{r}}$ and object pose  $_{\text{o}}\textbf{\textit{T}}$, 
the human can choose many different body configurations and poses to perform an operation. This raises the question of how the human will choose her configuation and pose, based on which we can compute the $\text{Comfort}$ value in Eq.~\ref{optimization}.

Given the robot configuration $\textbf{\textit{q}}_{\text{r}}$, object pose  $_{\text{o}}\textbf{\textit{T}}$ and the operation $(^\text{o}\textbf{\textit{f}},\, _{\text{t}}^{\text{o}}{\textbf{\textit{T}}})$, 
we define the set of all \textit{feasible} human pose and configuration combinations as
$Q_h\left(\textbf{\textit{q}}_{\text{r}},\,_{\text{o}}\textbf{\textit{T}},\,^\text{o}\textbf{\textit{f}},\, _{\text{t}}^{\text{o}}{\textbf{\textit{T}}}\right)$. 
By feasible, here we mean human poses and configurations that satisfy the stability constraints and 
the kinematic constraints (i.e., that puts the tool frame ${\sum}_{\text{t}}$ at the forceful operation point on the object as shown in Fig.~\ref{fig:problem_define}).
In other words, $Q_h$ includes the human poses and configurations the human can choose from to successfully perform the task.
Then, one can model the $\text{Comfort}$ using different schemes: 
\begin{itemize}
    \item One option is to take the average comfort value of all feasible human configurations and poses. Assuming $Q_h$ is a discretised set:
\begin{equation}\label{average_comfort}
\begin{split}
		\text{Comfort}\left(\textbf{\textit{q}}_{\text{r}},\,_{\text{o}}\textbf{\textit{T}},\,^\text{o}\textbf{\textit{f}},\, _{\text{t}}^{\text{o}}{\textbf{\textit{T}}}\right)=\\
        \underset{\textbf{\textit{q}}_{\text{h}},\,_{\text{h}}{\textbf{\textit{T}}} \in  Q_h}{\sum{\{w_{\text{M}}}}{\text{Muscular}\left(\textbf{\textit{q}}_{\text{h}}, _{\text{h}}{\textbf{\textit{T}}},  \, ^{\text{o}}{\textbf{\textit{f}}}, \, _{\text{t}}^{\text{o}}{\textbf{\textit{T}}} \right) }\\
{+w_{\text{P}}\text{Peripersonal}\left(\textbf{\textit{q}}_{\text{h}},\,\textbf{\textit{q}}_{\text{r}},\,_{\text{h}}{\textbf{\textit{T}}}\right)}\}/ | Q_h |
\end{split}
\end{equation}
Where $w_{\text{M}}$ and $w_{\text{P}}$ are weights. 
$| Q_h |$ indicates the size of $Q_h$. Note that the arguments to $Q_h$ is dropped above for clarity, but it is important to note that the set $Q_h$ is determined by the same arguments to Comfort function.

\item Another option is to assume the human will choose the most comfortable body configuration from among the feasible set: 
\begin{equation}\label{most}
\begin{split}
\text{Comfort}\left(\textbf{\textit{q}}_{\text{r}},\,_{\text{o}}\textbf{\textit{T}},\,^\text{o}\textbf{\textit{f}},\, _{\text{t}}^{\text{o}}{\textbf{\textit{T}}}\right)=\\
\underset{\textbf{\textit{q}}_{\text{h}},\,_{\text{h}}{\textbf{\textit{T}}} \in  Q_h }{\max\{w_{\text{M}}}{\text{Muscular}\left(\textbf{\textit{q}}_{\text{h}}, _{\text{h}}{\textbf{\textit{T}}},  \, ^{\text{o}}{\textbf{\textit{f}}}, \, _{\text{t}}^{\text{o}}{\textbf{\textit{T}}} \right) }\\
{+w_{\text{P}}\text{Peripersonal}\left(\textbf{\textit{q}}_{\text{h}},\,\textbf{\textit{q}}_{\text{r}},\,_{\text{h}}{\textbf{\textit{T}}}\right)}\}
\end{split}
\end{equation}

\end{itemize}

In the rest of this work we assume the human is optimal and therefore adopts the latter option, and we solve the optimization problem defined by Eq.~\ref{optimization} and~\ref{most}. The problem can alternatively also be solved for Eq.~\ref{optimization} and~\ref{average_comfort}.

\section{Muscular Comfort}
\label{sec:Muscular} 

This section introduces the metric used to quantify human muscular comfort, given a candidate configuration of the human-robot system and the forceful operation $\left(\textbf{\textit{q}}_{\text{h}}, \, _{\text{h}}{\textbf{\textit{T}}},  \, ^{\text{o}}{\textbf{\textit{f}}}, \, _{\text{t}}^{\text{o}}{\textbf{\textit{T}}}, \, _{\text{o}}{\textbf{\textit{T}}} \right)$.
This is a crucial element of the proposed system since the planner needs a
consistent quantitative assessment of human comfort to be optimized. 

To do this we model the human arm as a serial-link kinematic chain with seven
degrees of freedom (DOFs) \textemdash two spherical and one revolute joint
respectively at the shoulder, wrist and elbow \footnote{An important limitation
of this current formulation is that it only considers the human arm in the
comfort metric, and ignores the rest of the body. It is our intention to extend
the formulation and our planner to the whole body in the future.}.  The
configuration of this seven-DOF human arm is represented with the vector
$\myvec q _\human $. We use $\mymatrix{J}_\human ( \myvec q _\human )$ to
represent the geometric Jacobian of the human configuration (which we simplify
to $\mymatrix{J}_\human$ dropping the argument in this section). Furthermore,
in this section, we use $\myvec{f}$ to refer to the force the human is applying
in the world frame. Given the object pose $_\obj\mymatrix{T}$, the tool pose
$_{\text{t}}^{\text{o}}{\textbf{\textit{T}}}$, and the force in the object
frame $^\obj\myvec{f}$, $\myvec{f}$ can be found by a norm-preserving
transformation.

Assuming quasi-static movements, the effects of higher order dynamics can be
neglected and the human joint torques depend solely on the exerted force
$\myvec{f}$ and the gravity effects.  We assume the gravitational forces
results from the contribution of the upper arm and forearm weights and centre
of mass, and the tool mass and pose.  Naturally, the gravitational forces are
well-defined in the inertial/world frame and do not depend on the arm
configuration. However their gravitational torque---respectively, over the
shoulder, shoulder and elbow system and for the full arm---depends on the joint
configuration.  We use $\rho_\human(\myvec q_\human)$ to refer to these
generalized gravitational forces for a given human arm configuration, and again
we drop the argument $\myvec q_\human$ in the rest of the section for clarity.

Then, for a given forceful operation and human configuration, the required torques at
the human joints can be computed using:
\begin{equation}
\myvec{\tau}_\human  =  \mymatrix J^{T}_\human ( \myvec \forcef + \myvec \rho_\human)
\label{eq:Torque Human}
\end{equation}
where $\myvec{\tau}_\human$ is the vector of torques at the human joints.

If a human arm worked exactly like a robotic manipulator, we could compare the
torque values in $\myvec{\tau}_\human$ with the torque limits of the human
joints
and we could devise a comfort metric based on how close the torque
values are to the limits; i.e. the further from the limits, the more
comfortable it would be.

However, a better modeling of how the human muscles work is presented in Tanaka
et al. \cite{2015_Tanaka_Nishikawa_Yamada_Tsuji__TransHaptics}, which we use to
build our comfort metric. There are two key differences in this model, compared
to a robotic manipulator model. When approximating human's muscular structure with a kinematic structure like ours:
\begin{itemize}
    \item The torque limits must depend on the configuration, i.e. the torque limits do not stay constant at different configurations as they do for a robot manipulator;
    \item The torque limits must depend on the direction of motion, e.g. the limits of the elbow must be modeled differently if it is flexing versus if it is extending.
\end{itemize}
This suggests that, for the human arm, we model the torque limits as a function
of the configuration and the differential change in the configuration: $\Pi (\myvec q_\human, \dot{\myvec q}_\human )$, which we represent 
as a diagonal matrix with human maximum torque elements, that is,
$\text{diag} \{\tau_{1}^{max}(\myvec q_\human, \dot{\myvec q}_\human ),\dots,\tau_{7}^{max}(\myvec q_\human, \dot{\myvec q}_\human ) \}$. 
The matrix $\mymatrix \Pi$
represents the nonlinear relationship between human torque limits and joints;
and each maximum torque element, $\tau_{i}^{max}$, is a nonlinear
function of the joint values and their directions. 
%
Experimental data to compute the values for $ \mymatrix \Pi (\myvec q_\human,
\dot{\myvec q}_\human ) $ for different human arm configurations can be found
in the excel work of Tanaka et al.
\cite{2015_Tanaka_Nishikawa_Yamada_Tsuji__TransHaptics} and the values therein
are used throughout this work. 

This formulation requires us to know $\dot{\myvec q}_\human$ to extract the
direction each human joint will be moving during the forceful operation. We
compute $\dot{\myvec q}_\human$ by assuming that the human follows a
Jacobian-transpose controller: ${\myvec q}_\human=\mymatrix J_\human^T(\myvec
q_\human) \dot{\myvec x}_\human$, where $\dot{\myvec x}_\human$ is the
human hand motion during the application of the forceful operation. In this
work, we assume the hand moves in the same direction with the operational
force.

Equipped with  $\mymatrix \Pi (\myvec q_\human, \dot{\myvec q}_\human ) $, 
instead of Eq.~\ref{eq:Torque Human}, we can compute the 
human joint torque activation levels $\myvec{\alpha}_{\human}=[\alpha_{1},\dots,\alpha_{7}]^{T}$:
\begin{equation}
    \myvec{\alpha}_{\human}  =  \mymatrix \Pi^{-1} \mymatrix J^{T}_\human ( \myvec \forcef + \myvec \rho_\human)
\label{eq:Activation Level Human}
\end{equation}
where $\alpha_{i}\in[0,1]$ represents the activation ratio of the
$i$-th joint's torque $\tau_{i}$ to its maximum torque $\tau_{i}^{max}$---under maximum voluntary contraction and taking the 
muscle tension is nearly proportional to the muscle activation level. 

In this work, we use the norm of the activation level vector to compute the muscular human comfort. Specifically, we use
\begin{equation}
    \text{Muscular} ( \myvec q _\human, { _{\obj} \mymatrix{T} },  {^\obj\myvec{f}} , { ^\obj _{\tool} \mymatrix{T} } ,  { _{\obj} \mymatrix{T} } ) = 1 / ||\myvec{\alpha}_{\human}||^2
\end{equation}

\section{Peripersonal-space comfort}
\label{sec:Peripersonal} 

In this section, we define our metric quantifying the human's
peripersonal-space comfort for human-robot cooperative forceful manipulation.
We define this metric so that the optimization problem, presented in
Sec.~\ref{sec:defi}, can find solutions for which the human and robot body are
positioned at a comfortable distance from each other during collaboration.

%

In particular, for peripersonal comfort and safety, one would not want the human and
robot body to get too close to each other during cooperation. Therefore, we
define the peripersonal space comfort in terms of the distance
between the human and robot bodies. Given a human configuration,
$\textbf{\textit{q}}_{\text{h}}$, and pose, $_{\text{h}}{\textbf{\textit{T}}}$,
    we represent the set of points on the human body as
    $P_\text{h}(\textbf{\textit{q}}_{\text{h}},_{\text{h}}{\textbf{\textit{T}}})=\left\{p_\text{h}^1, p_\text{h}^2,...,p_\text{h}^n\right\}$
    where $p_\text{h}^i \in \mathbb{R}^3$.  Similarly, given a robot configuration
    $\textbf{\textit{q}}_{\text{r}}$, we represent the set of points on the
    robot body as $P_\text{r}(\textbf{\textit{q}}_{\text{r}})=\left\{p_\text{r}^1,
    p_\text{r}^2,...,p_\text{r}^m\right\}$  where $p_\text{r}^i \in \mathbb{R}^3$.
    (For clarity, in the rest of this work, we will simply use $P_\text{h}$ and $P_\text{r}$, dropping the arguments specifying the human/robot configurations and poses.)
Then, one can define the peripersonal comfort in terms of the distances
between these two set of points.

One option is to consider the minimum distance between the human and the robot,
and hypothesize the cooperation would be as comfortable as the minimum distance allows:
\begin{equation}\label{eq:peri_min_distance}
\text{Peripersonal}\left(\textbf{\textit{q}}_{\text{h}},\,\textbf{\textit{q}}_{\text{r}},\,_{\text{h}}{\textbf{\textit{T}}}\right)=
\min_{\forall p_\text{h} \in P_\text{h}, \forall p_\text{r} \in P_\text{r}} || p_\text{h} - p_\text{r} ||
\end{equation}
Then, the optimization process in Sec.~\ref{sec:defi} would be maximizing this minimum distance between the human and the robot.

While this metric may work well for different types of HRI,
it does not work well in the scenarios discussed in this work, since the
minimum distance between the human and the robot have a tight upper bound for
our tasks.  As shown in Fig.~\ref{fig:problem_define} the grippers of the robot
and the human hand on the tool are naturally the closest points. Thus,
maximizing the minimum distance will mostly only maximize the distance between
human hand and robot grippers, completely ignoring any other part of the body.

To address this issue and to take all points into account, one can instead consider the average distance over all points in the human body:
\begin{gather}\label{eq:dis_summin}
\text{Peripersonal}\left(\textbf{\textit{q}}_{\text{h}},\,\textbf{\textit{q}}_{\text{r}},\,_{\text{h}}{\textbf{\textit{T}}}\right)=
\frac{1}{|P_\text{h}|}\sum_{p_\text{h} \in P_\text{h}} \min_{\forall p_\text{r} \in P_\text{r}} || p_\text{h} - p_\text{r} ||
\end{gather}
While this metric takes all human body points into account, it also has a
problem: the large distances dominate this metric.  If a human configuration
has one point that is very far away from the robot but many other points that
are very close to the robot, then, 
this metric may provide misleading results.

This suggests we need a metric that takes all human body points into account,
but that also gives more weight to points with small distances to the robot and less
weight to points with large distances to the robot. Therefore, we propose the peripersonal space metric:
\begin{gather}\label{eq:dis_weighted}
\text{Peripersonal}\left(\textbf{\textit{q}}_{\text{h}},\,\textbf{\textit{q}}_{\text{r}},\,_{\text{h}}{\textbf{\textit{T}}}\right)=
\frac{1}{|P_\text{h}|}\sum_{p_\text{h} \in P_\text{h}} w(p_\text{h}) \min_{\forall p_\text{r} \in P_\text{r}} || p_\text{h} - p_\text{r} ||
\end{gather}
where $w(p_\text{h})$ is the weight of the point $p_\text{h}$ defined as
\begin{equation}\label{weight}
    w(p_\text{h})
    =1-\frac{\min_{\forall p_\text{r} \in P_\text{r}} || p_\text{h} - p_\text{r} ||}{\sum_{p_\text{h} \in P_\text{h}} \min_{\forall p_\text{r} \in P_\text{r}} || p_\text{h} - p_\text{r} ||}
\end{equation}
It associates a small distance with a large weight, thus controlling all distances efficiently.
Eq.~\ref{eq:dis_weighted}~and~\ref{weight} together define the peripersonal comfort metric used in this work.

\section{Stability Checking}
\label{sec:Feasibility} 

In addition to optimizing the muscular and peripersonal space comfort, the
optimization problem presented in Eq.~\ref{optimization} also makes sure that
the robot configuration, $\textbf{\textit{q}}_{\text{r}}$, the object pose,
$_{\text{o}}\textbf{\textit{T}}$, and the grasp, $\textit{\textbf{g}}$, are
chosen such that the object and the robot are stable during the application of the
force on the object by the human. This constraint is represented by the
Is\_stable function in Eq.~\ref{optimization}.  This function validates that:
\begin{itemize}
    \item The torque distribution on the robot arms' joints will not exceed the torque limits; 
    \item The grip forces, e.g. the frictional forces between robot fingers, will be able to resist the external force applied on the object.
\end{itemize}

Due to space limitation we skip the mathematical formulation
here, which can be found in our previous work
\cite{chen2018manipulation}, in
Section III-A-1, titled \textit{Stability Check}.

\section{Experiments and results}
\label{sec:Experiments} 

We implemented the comfort metrics, the stability check, and the optimization
process in Python.  We used the Baxter robot from Rethink Robotics and OpenRAVE (openrave.org)
for forward/inverse-kinematic calculations for the
human and robot model.  We used the SciPy library for optimization,
particularly the SLSQP
implementation\footnote{https://docs.scipy.org/doc/scipy/reference/optimize.minimize-slsqp.html}.
We used 64 points distributed around the whole body model of the human and 56
points for the whole body of the robot to compute the peripersonal-space
metric. We weighted the muscular and peripersonal space comfort equally during the optimization, 
yet it is worth noting that different weighting schemes may be chosen accordinglying to a given task and human preferences, e.g., their priors on HRI.

We consider two tasks. First, the task of cutting a circle out of a foam board.
We divided the circle into 16 discrete points
(Fig.~\ref{fig:comfortcurveData}). We call the application of the cutting force
on one of these 16 points a cutting operation; i.e. the discrete points along
the circle give us 16 different cutting operations.
The force
direction was aligned with the cutting direction. Using a force-torque sensor
we estimated the force required for cutting into foam as 30N. The second task
is the drilling task. 
We identified 9 points uniformly distributed on a foam
board in a 3-by-3 grid. 
We call the application of the drilling force on one of these 9 points a drilling operation.
The force direction was taken normal and into the
board. Using a force-torque sensor we estimated the force for drilling into the
foam as 15N.

\subsection{Analysis of the Muscular Comfort Metric}

First, to better understand and validate the proposed muscular comfort metric
we compared its predictions with known comfortable and uncomfortable configurations. 

Take for example, the task of cutting off a circular piece of a board, as shown in Figs.~\ref{fig:subfig:cut5good} and \ref{fig:subfig:cut5bad}. 
From the task definition, we expect fluctuations of muscular comfort at different configurations---from comfortable to very uncomfortable ones.  
Particularly, configurations at the proximity of joint limits, specially at the wrist spherical joint, 
are known to be very uncomfortable and have less force generation capabilities. 

\begin{figure}[!t]
	\centering
	\includegraphics[width=3.4in, angle=-0]{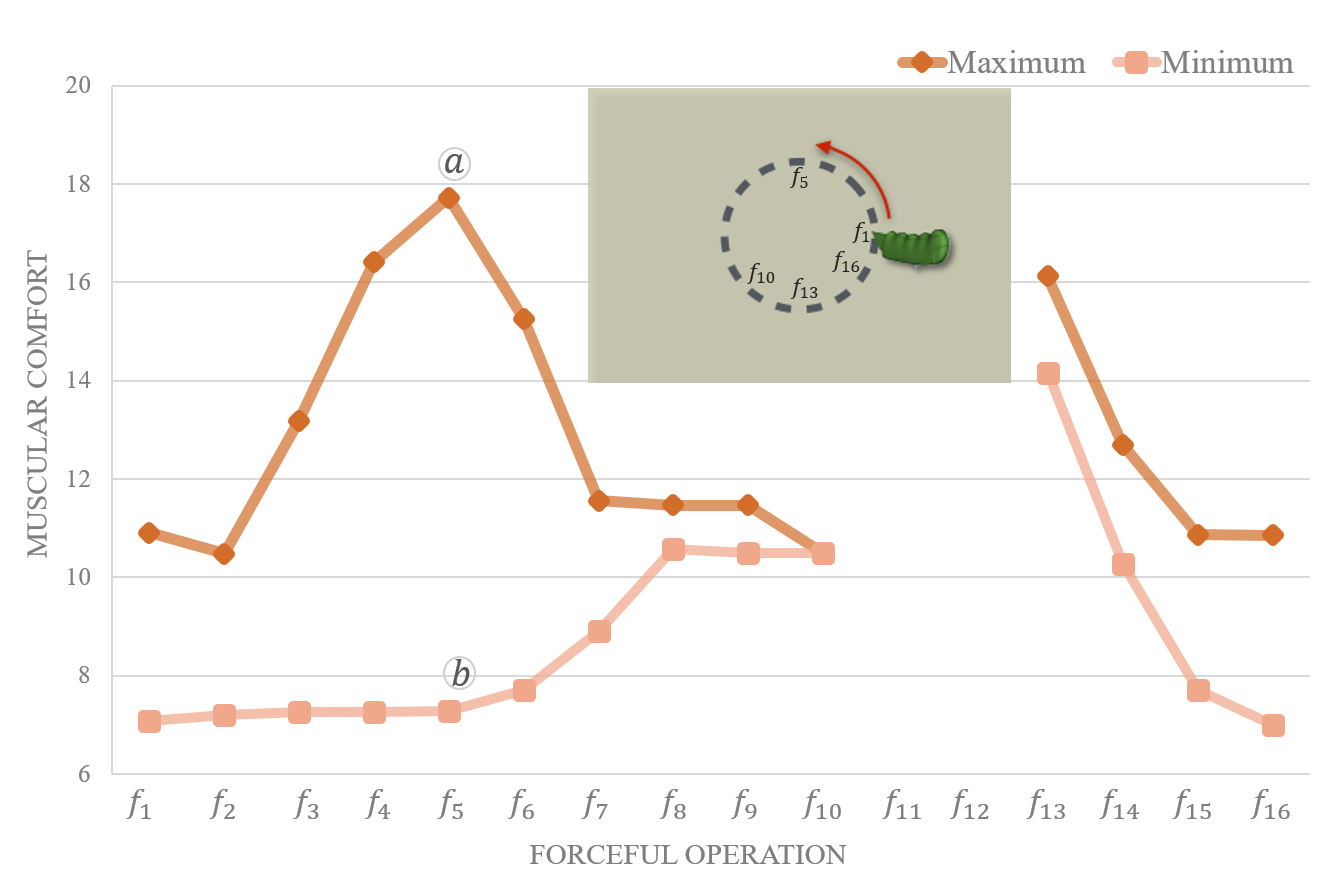}
    \caption{Muscular comfort values for different cutting points along a circle. The circled letters on the graph correspond to the configurations in Fig.~\ref{fig:comfortcurve}.
    }
	\label{fig:comfortcurveData}
\end{figure}

For each of the 16 discretized points along the circle, we computed a set of
feasible inverse-kinematics (IK) solutions for the human to perform the
forceful operation at that point. Then we computed the corresponding muscular
comfort values for all the IK solutions.
We present the computed muscular comfort values in Fig.~\ref{fig:comfortcurveData}. 
The dark orange curve depicts the change of muscular comfort when human is at the most comfortable configuration for a cutting operation, whilst the light orange depicts the worst (most uncomfortable) configuration for that cutting point. 
The range of possible human choices, illustrated in Fig.~\ref{fig:comfortcurve}, coincides with the human arm redundancy stressed in Sec.~\ref{sec:defi}. 
For the two possible configurations in Fig.~\ref{fig:comfortcurve}, our experience is that configuration \textcircled{{\small a}} is much more comfortable than configuration \textcircled{{\small b}}, which coincides with the results based on the muscular comfort metric.

\begin{figure}[!t]
	\centering
	\mbox{
		\subfigure[config.{\small  \textcircled{a}}]{{
				\label{fig:subfig:cut5good}  
				\includegraphics[width=1.3 in, angle=-0]{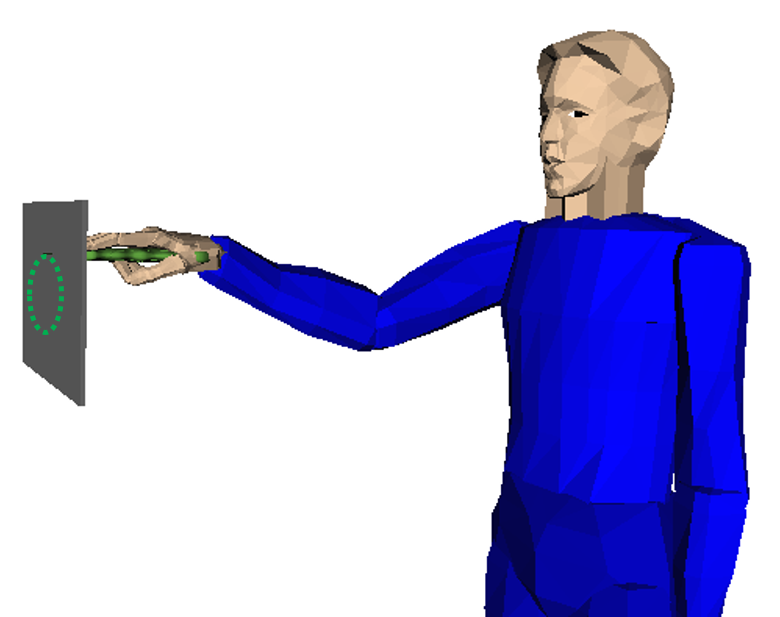}}}
		
		\subfigure[config.{\small  \textcircled{b}}]{{
				\label{fig:subfig:cut5bad}  
				\includegraphics[width=1.3 in, angle=-0]{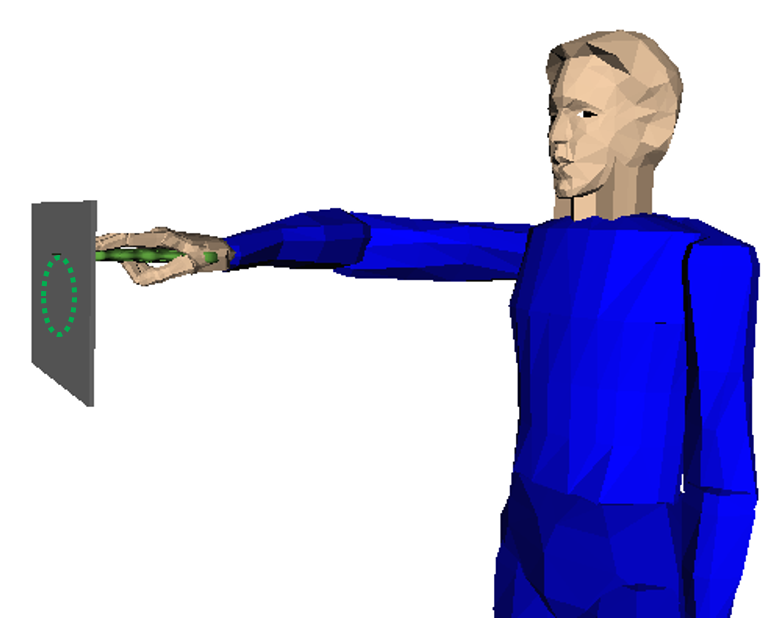}}}	
	}
    \caption{Two different arm postures while cutting the top of the board with different muscle comfort performance (see force position $f_5$ in Fig.~\ref{fig:comfortcurveData} with muscular comfort values {\small  \textcircled{a}} and {\small  \textcircled{b}}). }
	\label{fig:comfortcurve}
\end{figure}


At the proximity of joint limits shown in force position $f_{10}$ (positions $f_{11}$
and $f_{12}$ are not feasible due to joint limits), we can see a degradation in muscular
comfort and that the range of possible comfort values is narrower (due to the possible loss of a DoF). 
Finally, we can also see consistency in result at
the point of circular closing (force positions 16 to 1)

Fig.~\ref{fig:comfortcurveData} also shows us that it is critical to plan
optimal robot and object configurations to perform comfortable human-robot
collaboration. For example, for the circular-cutting task, if a robot keeps
rotating the object to enable the human configuration in
Fig.~\ref{fig:subfig:cut5good}, the required muscular effort for the human can
stay low during the cutting of the whole circle.

%
%

\subsection{Optimization performance: Simulation}

We evaluated the performance of the combined optimization of our comfort metrics by comparing it to other planners.
We implemented four planners:
\begin{itemize}
    \item{Comfort planner}: Our planner optimizing both metrics.
    \item{Random planner}: Given a forceful operation, the random planner randomly picks an object pose within the reachable space of the robot. Then it searches one feasible robot configuration in terms of static stability, and then chooses the optimized human configuration in terms of both muscular and peripersonal comfort under the constraint of human reachability to perform the task. The planner acts as baseline.
    \item{Muscular planner}: The muscular planner only takes the muscular comfort as cost (i.e., $w_\text{P}=0$ in Eq.~\ref{most}).
    \item{Peripersonal planner}: Similarly, the planner just maximizes the human's peripersonal comfort  (i.e., $w_\text{M}=0$).
\end{itemize}

\begin{table*}[]
	\centering
\caption{Average comfort results of four planners on 16 cutting and 9 drilling operations. Normalized with Random Planner results. 
Larger value is more comfort.} 
	\label{table:average}
	\begin{tabular}{@{}ccccccccc@{}}
		\toprule
		& \multicolumn{2}{c}{Random Planner} & \multicolumn{2}{c}{Comfort Planner}  & \multicolumn{2}{c}{Peripersonal Planner} & \multicolumn{2}{c}{Muscular Planner} \\ \cmidrule(l){2-3} \cmidrule(l){4-5} \cmidrule(l){6-7} \cmidrule(l){8-9} 
		& Musc. Comf.      & Perip. Comf.    & Musc. Comf.    & Perip. Comf.     & Musc. Comf.       & Perip. Comf.       & Musc. Comf.     & Perip. Comf.     \\ \midrule
		Cutting  & 1             & 1           & 11.19  & 2.08 & 0.34  & \textbf{2.09 }& \textbf{15.66}  & 1.02   \\
		Drilling & 1             & 1           & 19.24 & \textbf{2.55} & 5.78   & 2.51 & \textbf{22.45 } & 0.94  \\ \bottomrule
	\end{tabular}
\end{table*}

We ran all our planners on the 25 operations (16 cutting and 9 drilling).
Table~\ref{table:average} shows the average results of the four planners for both cutting and drilling operations.
To improve readability, results from the random planner were used as baseline to normalize the results of other planners:
For each task (drilling and cutting), muscular comfort values are normalized using the muscular comfort value of Random Planner. Similarly for the peripersonal-space comfort values.

As expected, the Peripersonal Planner outperforms others for peripersonal comfort (improvement of $2.09$ times over the random planner), yet it performs poorly ($0.34$ of the random planner) when it comes to muscular comfort. 
The muscular planner has similar performance: it outperforms for muscular comfort but performs poorly when considering peripersonal comfort. 
In constrast to other planners, the proposed Comfort Planner is able to optimize both metrics close to the values achieved by the planners optimizing individual metrics.
This highlights that it is possible to cope with both muscular and peripersonal comfort without much compromise.

%
%

\begin{figure*}[!thpb]
	\begin{center}
		\mbox{
			
            \subfigure[Random Planner]{{
					\label{fig:subfig:random-cut}  
					\includegraphics[width=1.6 in, angle=-0]{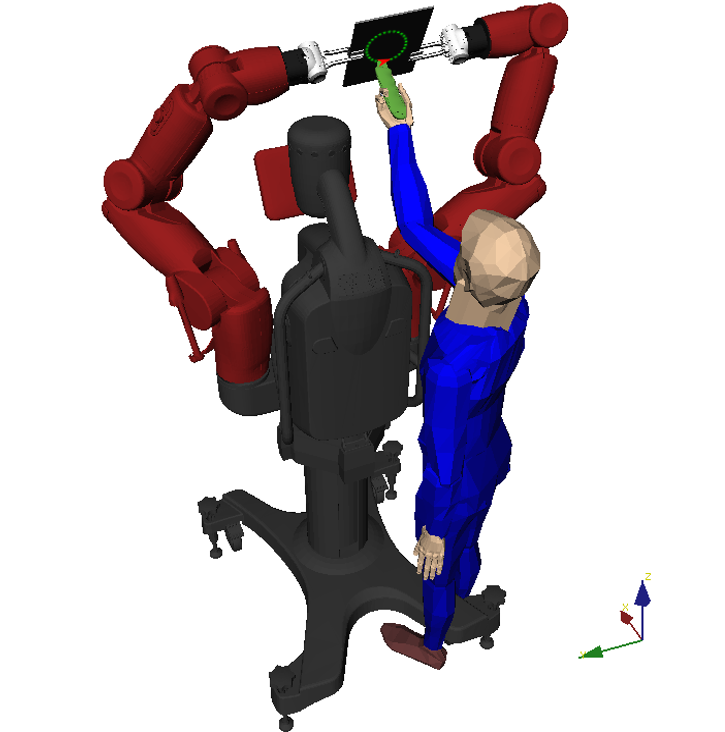}}}
				
                    \subfigure[Comfort planner]{{
		\label{fig:subfig:comfort-cut}  
		\includegraphics[width=1.6 in, angle=-0]{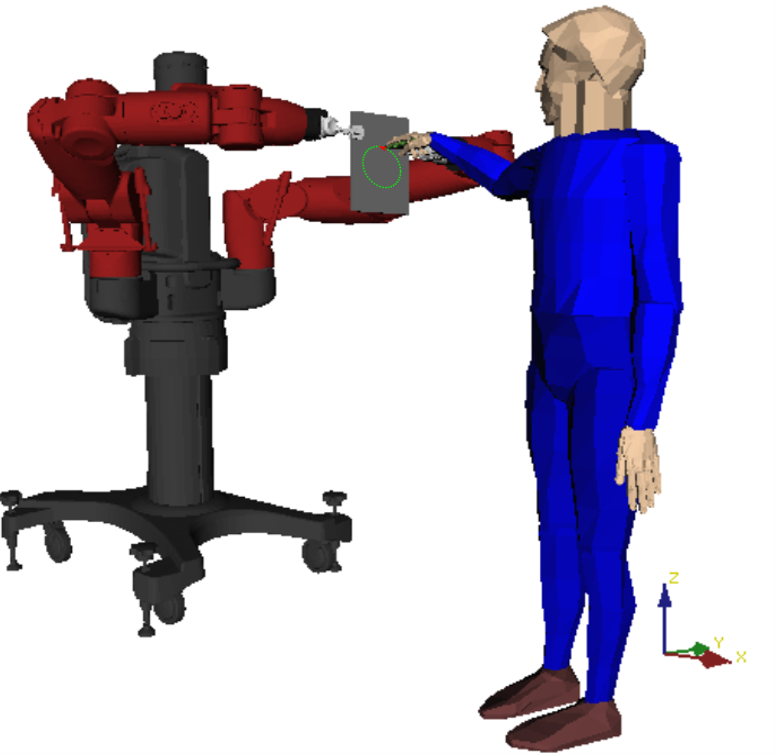}}}								
        \subfigure[Muscular planner]{{
					\label{fig:subfig:muscular-cut}  
					\includegraphics[width=1.6 in, angle=-0]{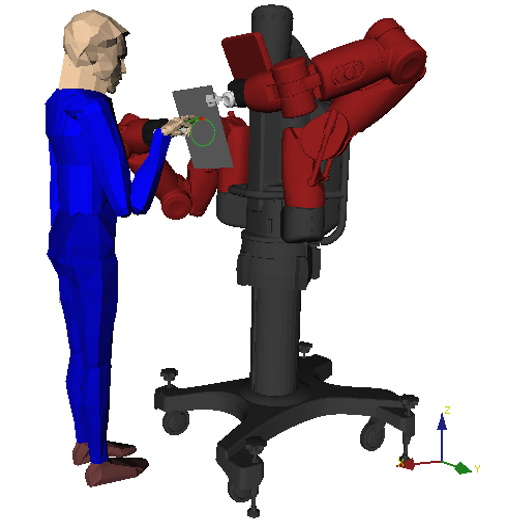}}}		
			
                    \subfigure[Peripersonal planner]{{
					\label{fig:subfig:safety-cut}  
					\includegraphics[width=1.6 in, angle=-0]{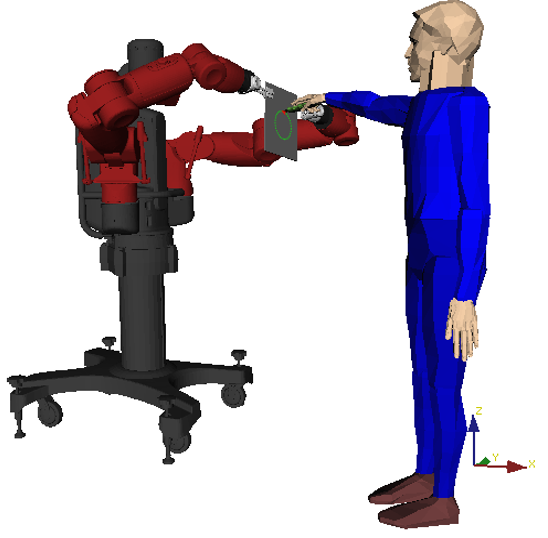}}}						

		}
		
        \caption{Optimization results for a cutting operation. The comfort values are: 
        (a) Musc. comfort: 8.34; Perip. comfort: 20.43.
        (b) Musc. comfort: 47.10; Perip. comfort: 43.30.
        (c) Musc. comfort: 48.53; Perip. comfort: 23.38.
        (d) Musc. comfort: 3.98; Perip. comfort: 44.23.
         }
		\label{fig:cut_optimizationresults}
	\end{center}
    \vspace{-3mm}
\end{figure*}

\begin{figure*}[!thpb]
	\begin{center}
		\mbox{
            \subfigure[Random planner]{{
					\label{fig:subfig:random-drill} 
					\includegraphics[width=1.65 in, angle=-0]{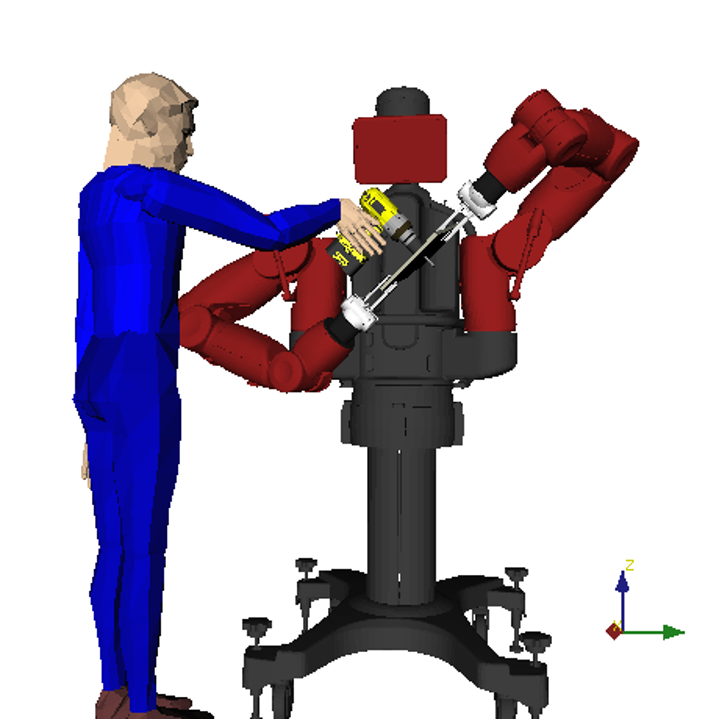}}}
                    \subfigure[Comfort planner]{{
		\label{fig:subfig:comfort-drill}  
		\includegraphics[width=1.6 in, angle=-0]{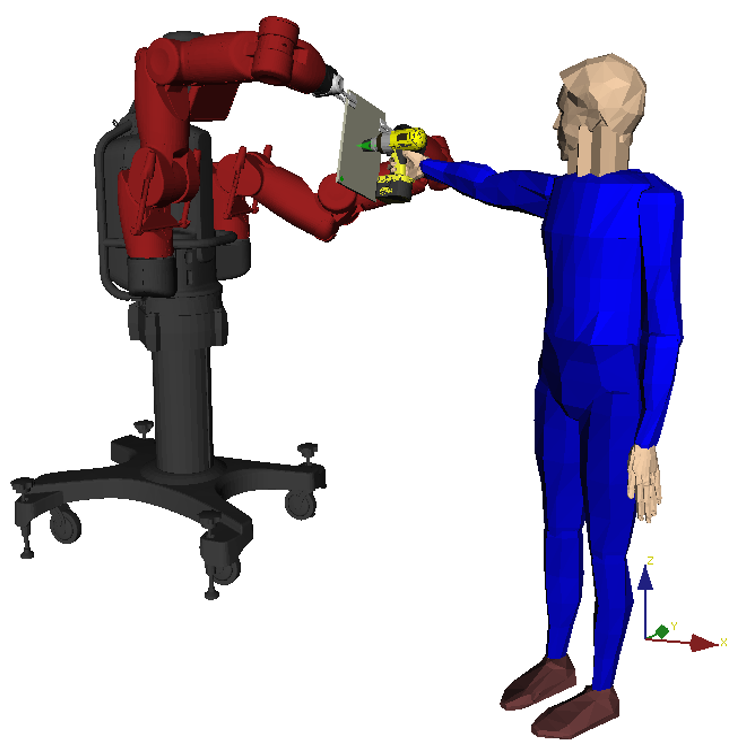}}}	

        \subfigure[Muscular planner]{{
					\label{fig:subfig:muscular-drill}  
					\includegraphics[width=1.6 in, angle=-0]{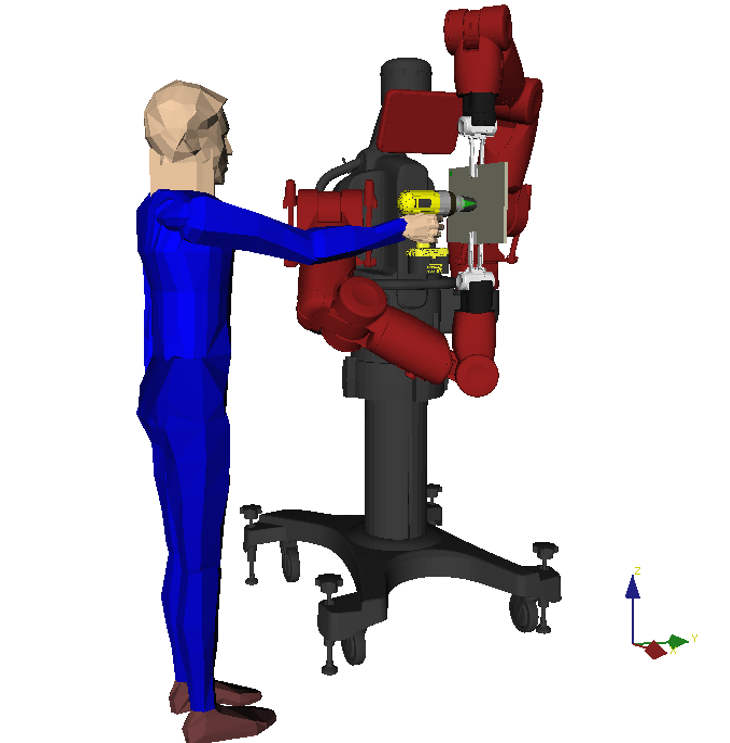}}}		
			
                    \subfigure[Peripersonal planner]{{
					\label{fig:subfig:safety-drill}  
					\includegraphics[width=1.6 in, angle=-0]{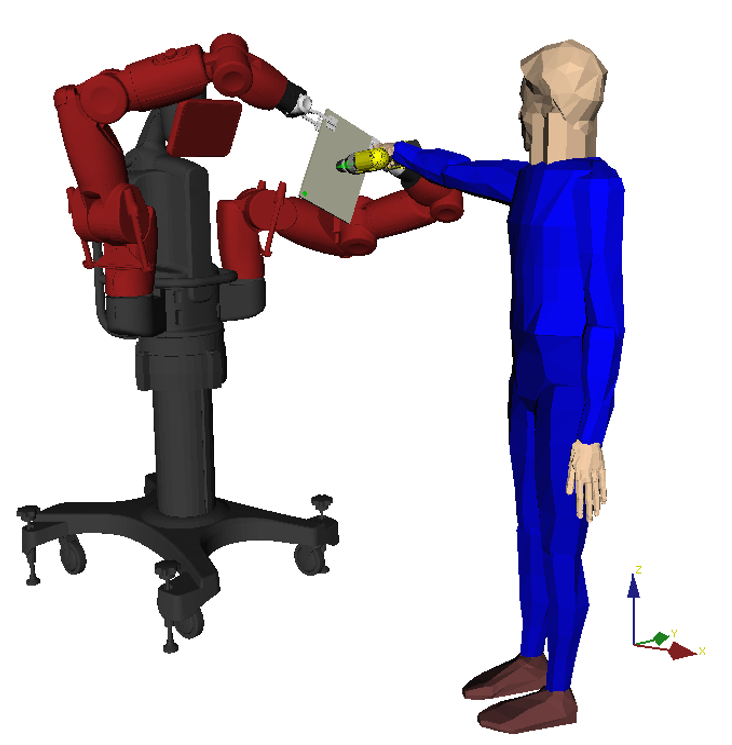}}}					
		
		}		
		\caption{Optimization results for a drilling operation. The comfort values are:
        (a) Musc. comfort: 3.70; Perip. comfort: 20.21.
        (b) Musc. comfort: 59.19; Perip. comfort: 44.75.
        (c) Musc. comfort: 42.33; Perip. comfort: 21.88.
        (d) Musc. comfort: 4.72; Perip. comfort: 44.40.
    }
		\label{fig:drill_optimizationresults}
	\end{center}
\end{figure*}

We present configurations found by the planners for one cutting operation
in Fig.~\ref{fig:cut_optimizationresults} and for one drilling operation in
Fig.~\ref{fig:drill_optimizationresults}. Again, we see that the Comfort
Planner can generate solutions that are comfortable according to
both metrics.

We realized that a limitation of the proposed implementation is the time it takes to generate results.
To compute one configuration, the Comfort Planner took 9.4 seconds on average over the 25 operations, with a standard deviation of
1.8. Most of this time was spent on generating
IK solutions. These time results indicate that faster schemes
must be developed if the robot is to perform fluent interaction with a human
during a continuous task. We are currently investigating methods based on
precomputing IK solutions.

\subsection{Human experiments}

Finally, 
we conducted a series of human-robot experiments to evaluate human experience and comfort perception 
during a forceful operation with real cooperative robot. 

The Baxter robot 
gravity compensation mode is activated by a sensor at the arms cuff on which participants could easily hold.
Using this feature, $5$ participants were asked to engage in motion activities with the Baxter robot in order to familiarize themselves with the robot and to improve their psychological comfort.
The participants were also briefed about the goal of the study and the tasks they were expected to perform. 

\begin{figure}[!t]
	\begin{center}
		\mbox{
			\subfigure[Random planner - Drilling]{{
					\label{fig:subfig:real:random-drill} 
					\includegraphics[width=1.6 in, angle=-0]{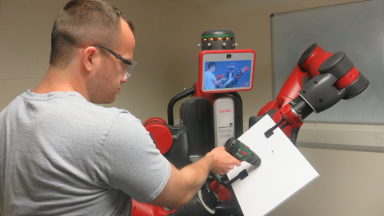}}}
			\subfigure[Random planner - Cutting]{{
		\label{fig:subfig:real:random-cut}  
		\includegraphics[width=1.6 in, angle=-0]{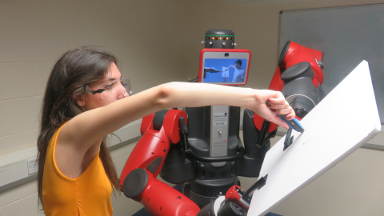}}}	
		} 
		\mbox{
			\subfigure[Comfort based planner - Drilling]{{
					\label{fig:subfig:real:comfort-drill}  
					\includegraphics[width=1.6 in, angle=-0]{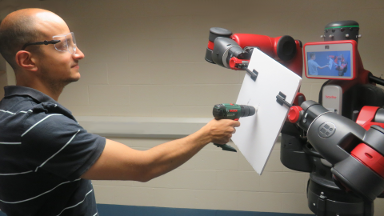}}}		
			
			\subfigure[Comfort based planner - Cutting]{{
					\label{fig:subfig:real:comfort-cut}  
					\includegraphics[width=1.6 in, angle=-0]{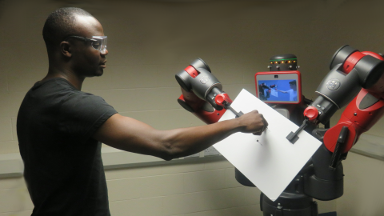}}}					
		
		}		
		\caption{Cooperative drilling and cutting experiments.}
		\label{fig:realExperiments}
	\end{center}
\end{figure}

The tasks required were three pairs of drilling and cutting on a foam board tightly held by the Baxter robot. 
First, (i) the participants were asked to move the robot arms using the gravity compensation mode
to a position they felt comfortable and safe to interact, and drill through the center of a foam board; 
then (ii) they were asked to do the same but for a cutting operation. 
We then moved the Baxter arms to previously set configurations which were randomly obtained (one for drilling and another one for cutting operations). 
Participants were then asked to (iii) drill and then (iv) cut the board at random robot configurations. 
Finally, 
we moved the Baxter arms to previously decided configurations 
based on our Comfort Planner 
(one optimized robot configuration for drilling and another one for cutting). 
Participants were then asked to (v) drill and (vi) cut at the optimized configurations. 
Some of these operations are shown in Fig.~\ref{fig:realExperiments}.

%

\begin{figure}[!t]
    \centering\includegraphics[width=0.9\columnwidth]{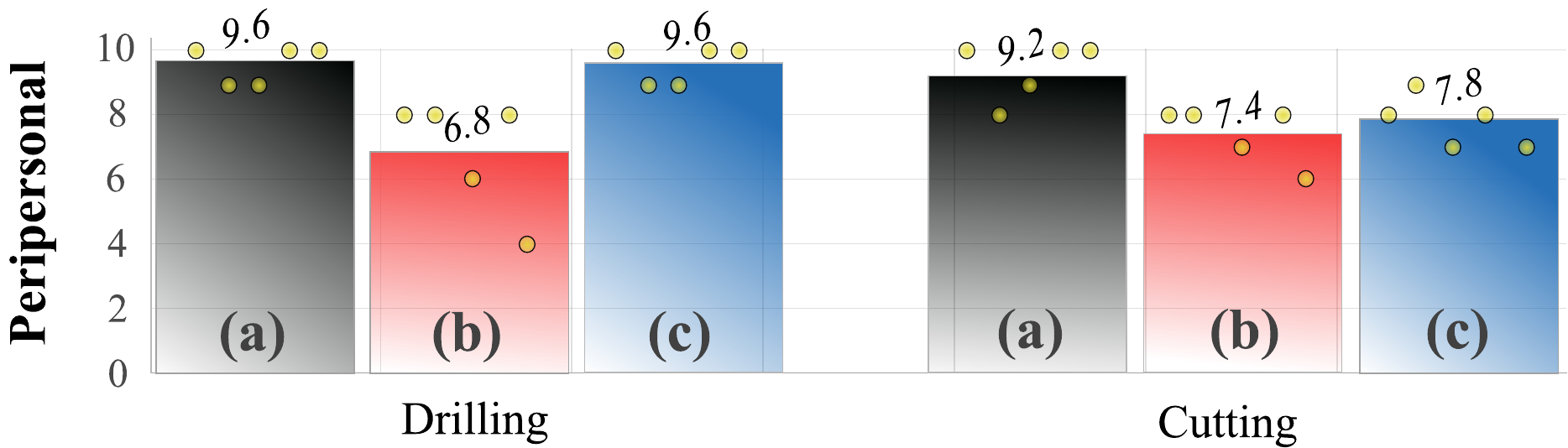}%
    \caption{Average peripersonal comfort perception from users (yellow dots) for drilling (left) and cutting (right) within the  \textbf{(a)} user defined object pose (black), \textbf{(b)} random planner (red), and \textbf{(c)} optimized pose (blue).}
\label{fig:baxter results:safety}
\end{figure}

\begin{figure}[!t]
    \centering\includegraphics[width=0.9\columnwidth]{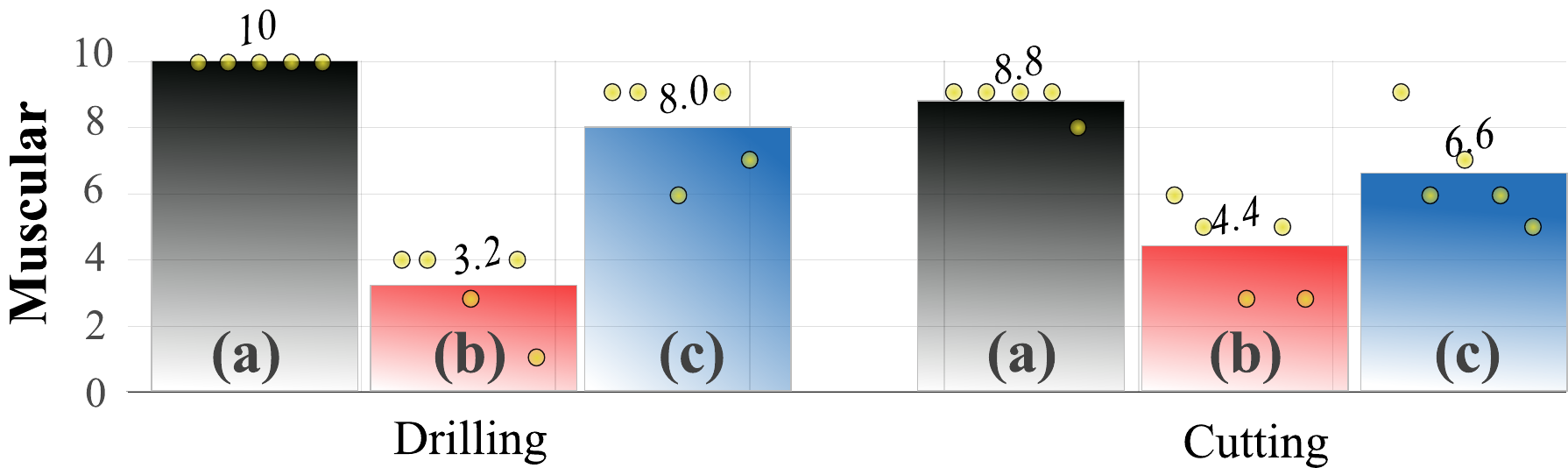}%
    \caption{Average muscular comfort perception from users  (yellow dots) for drilling (left) and cutting (right) within the  \textbf{(a)} user defined object pose (black), \textbf{(b)} random planner (red), and \textbf{(c)} optimized pose (blue).}
\label{fig:baxter results:comfort}
\end{figure}

After the experiments, for each scenario, 
participants were asked to grade from 1 to 10 (higher better), their perception on the human-robot safety and peripersonal comfort, and their perception over the muscular forces demanded during the task execution. 
The scores are summarized in Fig.~\ref{fig:baxter results:safety} for peripersonal comfort and in Fig.~\ref{fig:baxter results:comfort} for muscular comfort. 
From both figures, it is easy to see participants perception regarding the peripersonal and muscular comfort agrees with our expectations, particularly for the drilling operation. 
Cutting operation was clearly not as comfortable as drilling---even for participants' own choice of object positioning. 
Significant difference between the Random Planner and the Comfort Planner is seen for the muscular comfort perception for the drilling operation. 
With respect to the peripersonal space comfort, while the participants showed a preference for the Comfort Planner configurations compared to the Random configurations, the difference was not as significant. During the experiments some participants expressed that they did not prefer to be very far away from the robot, possible suggesting that, while the peripersonal distance must be maximized up to a certain degree, there may be a distance after which it starts to negatively affect the collaboration. 

\section{Conclusion and Future Work}

This work 
proposed a new technique for grasp planning and object positioning that explicitly concerns human configuration, safety and comfort. 
To this aim, we developed a new criteria to cope with both muscular comfort and safety perception 
in a quantiative manner which allowed us to analyze the desired comfort without having to rely on experience and intuition alone. 
For the muscular comfort, 
the proposed planner succed in minimizing the 
human joint-torque muscular activation level required to execute a given forceful task, 
whilst human safety perception was addressed through the optimization of the subject peripersonal space 
and distance during HRI. 
Both criteria were optimized taking human-robot kinematic constraints and ensuring safe and stable grasping from the robotic counterpart. 
Validity and effectiveness of our algorithm were first analyzed through different simulation scenarios and then 
validated with different human-robot experiments. The results highlighted the effectiveness of the proposed planner 
and the necessity to explicitly consider and plan for the human and not only for the task. 
Based on the results, our future work will focus on (i) improving the
peripersonal-space metric to better represent human perception, (ii) improving
the planning time to enable fluent human interaction on continuous forceful
tasks, and (iii) adding other metrics to our framework, e.g. metrics that take
into account the visibility of the task execution by the human.


%
%

%
%

\bibliographystyle{IEEEtran}
\bibliography{References/references,References/2018_ICRAWS_BIB,References/peripersonalDef}

\end{document}